\documentclass[usenames,dvipsnames,sigconf]{acmart}

\usepackage[nolist]{acronym}
\AtBeginDocument{%
 \providecommand\BibTeX{{%
    \normalfont B\kern-0.5em{\scshape i\kern-0.25em b}\kern-0.8em\TeX}}}

\usepackage{makecell}
\usepackage{enumitem}
\usepackage{multirow}
\usepackage{algorithm}
\usepackage{algpseudocode} 
\usepackage{tabularx}
\usepackage{hyperref}
\usepackage{amsmath}
\usepackage{float}
\usepackage{setspace}
\usepackage{multicol}
\usepackage{caption}
\usepackage{colortbl}
\definecolor{mygray}{gray}{.9}

\copyrightyear{2024}
\acmYear{2024}
\setcopyright{acmlicensed}\acmConference[MM '24]{Proceedings of the 32nd ACM International Conference on Multimedia}{October 28-November 1, 2024}{Melbourne, VIC, Australia}
\acmBooktitle{Proceedings of the 32nd ACM International Conference on Multimedia (MM '24), October 28-November 1, 2024, Melbourne, VIC, Australia}
\acmDOI{10.1145/3664647.3681402}
\acmISBN{979-8-4007-0686-8/24/10}

\begin{document}
\title{Dual Advancement of Representation Learning and Clustering for Sparse and Noisy Images} 

\author{Wenlin Li}
\authornote{Equal contribution.}
\email{wenlinli@stu.zuel.edu.cn}
\orcid{0009-0006-4366-1660}
\affiliation{%
  \institution{School of Statistics and Mathematics, Zhongnan University of Economics and Law}
  \city{Wuhan}
  \country{China}
  \postcode{430073}
}

\author{Yucheng Xu}
\orcid{0009-0004-9552-4305}
\authornotemark[1]
\email{yuchengxu@mail.nankai.edu.cn}
\affiliation{%
  \institution{School of Statistics and Data Science, Nankai University}
  \city{Tianjin}
  \country{China}
  \postcode{300071}
}

\author{Xiaoqing Zheng}
\orcid{0009-0001-5562-1176}
\email{xiaoqingzheng@stu.zuel.edu.cn}
\affiliation{%
  \institution{School of Statistics and Mathematics, Zhongnan University of Economics and Law}
  \city{Wuhan}
  \country{China}
  \postcode{430073}
}

\author{Suoya Han}
\orcid{0009-0004-8057-6727}
\email{suoyahan@stu.zuel.edu.cn}
\affiliation{%
  \institution{School of Statistics and Mathematics, Zhongnan University of Economics and Law}
  \city{Wuhan}
  \country{China}
}

\author{Jun Wang}
\orcid{0000-0002-9515-076X}
\email{jwang@iwudao.tech}
\affiliation{%
  \institution{iWudao}
  \city{Nanjing}
  \country{China}
  \postcode{430073}
}

\author{Xiaobo Sun}
\orcid{0000-0001-9876-5666}
\authornote{Corresponding author.}
\email{xsun@zuel.edu.cn}
\affiliation{%
  \institution{School of Statistics and Mathematics, Zhongnan University of Economics and Law}
  \city{Wuhan}
  \country{China}
  \postcode{430073}
}
\renewcommand{\shortauthors}{Wenlin Li et al.}

\begin{abstract}
    Sparse and noisy images (SNIs), like those  in spatial gene expression data, pose significant challenges for effective representation learning and clustering, which are essential for thorough data analysis and interpretation. In response to these challenges, we propose \textbf{D}ual \textbf{A}dvancement of \textbf{R}epresentation \textbf{L}earning and \textbf{C}lustering (\textit{\textbf{DARLC}}), an innovative framework that leverages contrastive learning to enhance the representations derived from masked image modeling. Simultaneously, \textit{DARLC} integrates cluster assignments in a cohesive, end-to-end approach. This integrated clustering strategy addresses the ``class collision problem'' inherent in contrastive learning, thus improving the quality of the resulting representations. To generate more plausible positive views for contrastive learning, we employ a graph attention network-based technique that produces denoised images as augmented data. As such, our framework offers a comprehensive approach that improves the learning of representations by enhancing their local perceptibility, distinctiveness, and the understanding of relational semantics. Furthermore, we utilize a Student's t mixture model to achieve more robust and adaptable clustering of SNIs. Extensive experiments, conducted across 12 different types of datasets consisting of SNIs, demonstrate that \textit{DARLC} surpasses the state-of-the-art methods in both image clustering and generating image representations that accurately capture gene interactions. Code is available at \href{https://github.com/zipging/DARLC}{https://github.com/zipging/DARLC}.
\end{abstract}




\maketitle
\section{Introduction}
\begin{figure}[t]
  \centering
  \includegraphics[width=\linewidth]{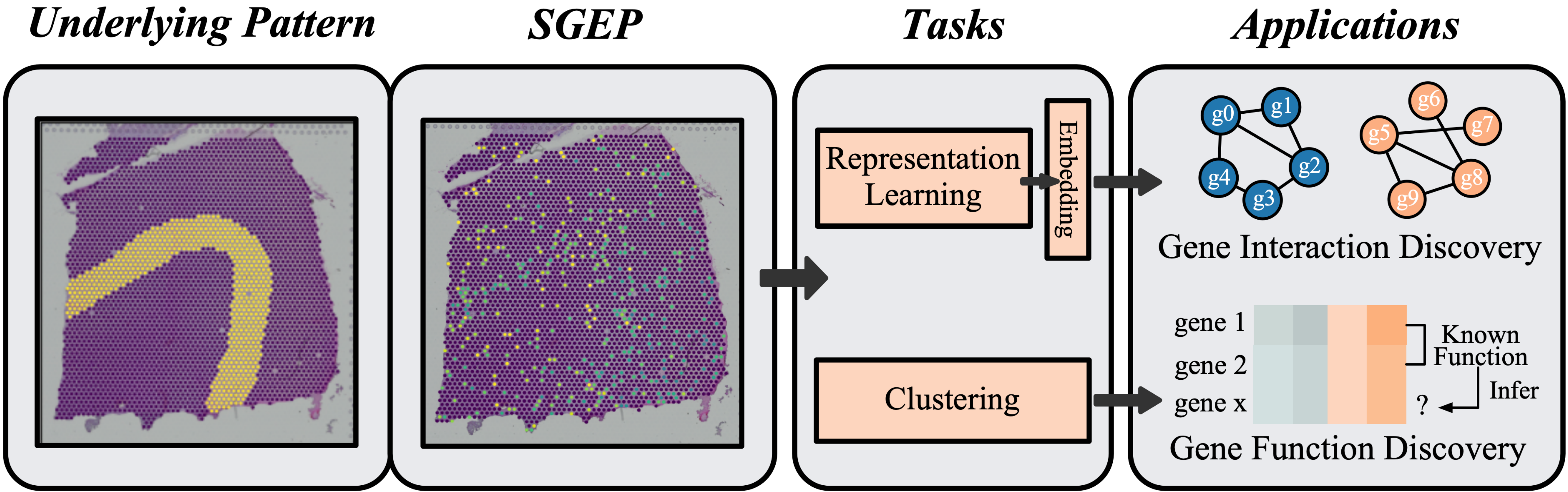}
  \caption{An example of sparse and noisy SGEP image is displayed in the second panel to the left. The gene expression levels across space are represented by pixel brightness, while void expression areas are displayed in purple. The first panel showcases the regions (cortical layer V) where the gene is significantly expressed, namely the major gene expression pattern. The right two panels illustrate the tasks and applications that utilize SGEP images.}
  \label{fig:Comparison between spase and normal}
\end{figure}

Sparse and noisy images (SNIs), commonly encountered in specialized fields like biomedical sciences, astronomy, and microscopy ~\cite{zhu2021spark,sloka2023image,wang2023genesegnet}, are characterized by extensive uninformative regions (e.g., voids or background areas), considerable image noise, and severely fragmented visual patterns. These characteristics significantly increase the complexity in analysis and interpretation. A prime example is spatial gene expression Pattern (SGEP) images generated through spatial transcriptomics (ST) technology~\cite{staahl2016visualization}. As illustrated in Figure \ref{fig:Comparison between spase and normal}, the high levels of sparsity and noise of an SGEP image complicate the discerning of its underlying major gene expression pattern.

Image clustering can group unlabelled images into distinct clusters, facilitating the exploration of image-implied semantics or functions. For example, clustering SGEP images offers a cost-effective means to identify groups of cofunctional genes and infer gene functions \cite{song2022detecting}. To obtain informative clustering results, it is essential to learn meaningful image representations, for which self-supervised learning (SSL) is the predominant approach in general scenarios. These include contrastive learning (CL) methods, exemplified by MoCo~\cite{he2020momentum}, and masked image modeling (MIM) such as MAE~\cite{he2022masked}. These methods offer distinct learning perspectives: MIM methods tend to learn local context-aware, holistic features for reconstruction tasks \cite{he2022masked}, while CL methods focus on learning instance-wise discriminative features \cite{huang2023contrastive}. Acknowledging the synergistic advantages of these methodologies, some researchers are endeavoring to utilize CL to refine the representations acquired through MIM~\cite{zhou2021ibot,huang2023contrastive}.  Moreover, many efforts~\cite{ren2022deep,xie2016unsupervised,li2021contrastive} are directed towards guiding the process of learning representations through clustering tasks. This is achieved by jointly learning representations and executing clustering in an integrated and end-to-end manner, yielding image representations that are well-suited for clustering tasks. 
 
However, for SNIs, both representation learning and clustering present significant challenges. Firstly, the widespread presence of uninformative voids or background areas, along with elevated noise levels and extremely fragmented visual patterns, substantially impedes the extraction of semantically meaningful visual features. This challenge has been highlighted in prior studies~\cite{lu2021sparse} and is further corroborated by our experiments (see Supplementary Table 1). Secondly, the inherent random noise across pixels induces considerable variability in visual patterns, even among images of the same category~\cite{song2022detecting}, exposing clustering algorithms to a high overfitting risk~\cite{askari2021fuzzy}.  

Inspired by the aforementioned works, in order to better analyze SNIs, we propose a novel and unified framework, named \textbf{D}ual \textbf{A}dvancement of \textbf{R}epresentation \textbf{L}earning and \textbf{C}lustering (\textit{\textbf{DARLC}}). This framework not only leverages CL to boost the representation learned by MIM but also jointly learns cluster assignments in a self-paced and end-to-end manner, further refining the representation. Nonetheless, our experiments (see Supplementary Table 2) showcase that conventional data augmentation techniques (e.g., cropping and rotating) are ineffective for SNIs, as the augmented images often contain substantial void regions and noise, hampering the extraction of informative visual features. To overcome this limitation, we introduce a data augmentation method based on a graph attention network (GAT)~\cite{velivckovic2017graph,dong2022deciphering} that aggregates information from neighboring pixels to enhance visual patterns, generating smoothed images that act as more plausible positive views so as to improve the effectiveness of contrastive learning. 

Additionally, we observed that the clustering algorithms used in many deep clustering methods are either sensitive to outliers, as demonstrated by the Gaussian mixture model (GMM) in DAGMM~\cite{zong2018deep} and manifold clustering in EDESC \cite{cai2022efficient}, or lack the flexibility to different
data distributions, such as the inflexible Cauchy kernel-based method in DEC ~\cite{xie2016unsupervised}. In response, \textit{DARLC} employs a specialized nonlinear projection head to normalize image embeddings, aligning them more closely with a t-distribution. This is followed by modeling with a Student's t mixture model (SMM) for soft clustering. SMM provides a more robust solution by down-weighing extreme values and is more adaptable by altering the degrees of freedom, making it particularly suitable for clustering in the context of SNIs. Furthermore, the clustering loss in \textit{DARLC} also serves to regularize CL, alleviating the ``class collision problem" that stems from false negative pairs in CL \cite{cai2020all}. Unlike existing regularized CL methods \cite{denize2023similarity,zheng2021weakly,li2020prototypical}, which directly integrate clustering into the CL throughout training, this clustering follows the ``warm-up" representation learning, significantly expediting training convergence and enhancing clustering accuracy, as demonstrated in our ablation study. All these features collectively contribute to the finding that \textit{DARLC}-generated image representations not only enhance clustering performance but also exhibit improvements in other semantic distance-based tasks, such as the discovery of functionally interactive genes. In summary, our main contributions are:
\begin{itemize}[left=0pt]
\setlength{\itemsep}{0.01pt}
    \item We propose \textit{DARLC}, a novel unified framework for dual advancement of representation learning and clustering for SNIs. \textit{DARLC} marks the first endeavor in integrating contrastive learning, MIM and deep clustering into a cohesive process for representation learning. The resultant representations enhance image clustering performance and benefit other semantic distance-based tasks. 
    \item \textit{DARLC} has developed a data augmentation method more suitable for SNIs, using a GAT to generate smoothed images as plausible positive views for CL. 
    \item An SMM-based method is designed to cluster SNIs in a more robust and adaptable manner. Additional features of this clustering method include a novel Laplacian loss for guiding the initial phase of clustering, and a differentiable cross-entropy hinge loss for controlling cluster sizes. This clustering also addresses the class collision problem by pulling close related instances.
    \item Extensive experiments have been conducted across 12 SNIs datasets. Our results show that \textit{DARLC} surpasses the state-of-the-art (SOTA) methods in both image clustering and generating image representations that can be effectively applied to specific downstream tasks.
\end{itemize}

\section{Related Works}
\begin{figure*}[t]
  \centering
  \includegraphics[width=\linewidth]{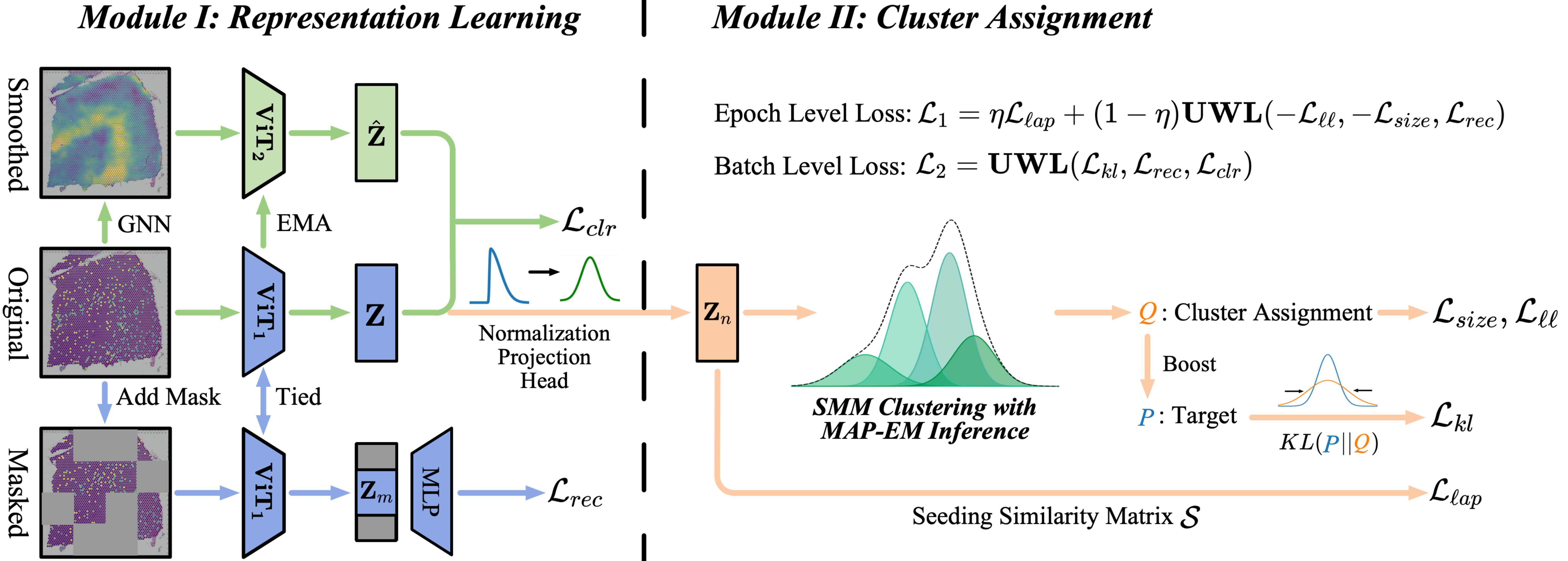}
  \caption{The framework of \textit{DARLC} consists of two components: the representation learning and the deep clustering. The representation learning component integrates  MIM and CL to generate image embeddings, which are then normalized through a non-linear projection head. Normalized representations are modeled by an SMM to derive their soft cluster assignments, which are used to construct various loss functions. With these loss functions, the two components are jointly optimized in a self-paced and end-to-end manner.} 
  \label{framework}
\end{figure*}
\subsection{Self-supervised Representation Learning for Images}
Most related SSL studies include CL and MIM methods. In CL, an input instance forms positive pairs with its augmented views, while forms negative pairs with other instances. Paradigmatic CL methods aim to learn instance discriminative representations by maximizing the similarity between positive pairs while minimizing it between negative pairs in a latent space \cite{chen2020simple,he2020momentum}. To address the class collision problem due to false negative samples, several studies \cite{li2020prototypical,denize2023similarity,zheng2021weakly} regularize CL with clustering, while
others 
\cite{grill2020bootstrap,caron2020unsupervised,caron2021emerging,chen2021exploring} bypass the using of negative samples altogether. In contrast, MIM methods focus on learning local context-aware features by restoring raw pixel values from masked image patches \cite{he2022masked,gao2022convmae,xie2022simmim,bao2021beit}. Several researchers are realizing the advantages of integrating these methodologies and endeavoring to utilize CL to refine MIM-generated representations~\cite{xie2022simmim,huang2023contrastive}. For instance, iBOT \cite{zhou2021ibot} contrasts between the reconstructed tokens of masked and unmasked image patches. Yet, to the best of our knowledge, \textit{DARLC} is the first method that learns image representations from all aspects of discriminability, local perceptability, and relational semantic structures. 

\subsection{Deep Image Clustering.}  

Related deep image clustering studies include deep autoencoder-based methods \cite{ren2022deep}, which couple representation learning with deep embedded clustering in an end-to-end manner, as exemplified by methods like DEC \cite{xie2016unsupervised,li2018discriminatively,guo2017improved}. Subsequent improvements to DEC focus on strategies like overweighing reliable samples (e.g., IDCEC ~\cite{lu2022improved} ), and replacing Euclidean distance-based clustering with deep subspace clustering (e.g., EDESC \cite{cai2022efficient}) or GMM-based clustering \cite{wang2021unsupervised,zong2018deep}. Recent studies, including CC \cite{li2021contrastive}, DCP \cite{lin2022dual}, CVCL \cite{chen2023deep}, and CCES~\cite{yin2023effective}, directly integrate CL into the clustering process by contrasting at both instance and cluster levels across views, generating a soft cluster assignment matrix as deep embeddings for iterative refinement. Compared to these methods, \textit{DARLC} offers a more comprehensive and potent mechanism for learning deep embeddings, a more robust and adaptable clustering algorithm, and a warm-up representation learning phase for accelerating clustering convergence.

\section{Methodology}

\subsection{Overview}

The framework of \textit{DARLC}, as illustrated in Figure \ref{framework} and Algorithm \ref{pseudo-code}, comprises two modules: a self-supervised representation learning and a deep clustering. The self-supervised representation learning module unifies CL and MIM, encompassing three encoders: an online encoder and a target momentum encoder for the contrastive branch, and a masked encoder for the MIM branch. All three encoders adopt the identical vision transformer (ViT) \cite{dosovitskiy2020image} architecture, with shared parameters between the online encoder and masked encoder. The parameters of the target momentum encoder are updated using exponential moving average (EMA), as suggested by BYOL \cite{grill2020bootstrap}. The initial phase of the unified representation learning involves a warm-up pretraining to generate preliminary embeddings, which are then normalized by a nonlinear projection head. This normalization aligns the embeddings more closely with t-distributions, setting the stage for subsequent t-distribution-based clustering. The deep clustering module utilizes a SMM to cluster the normalized embeddings, generating soft cluster assignment scores involved in the calculation of various loss functions. There are two types of loss functions: $\mathcal{L}_1$, an epoch-level loss function for maximizing the empirical likelihood of observed instances, and $\mathcal{L}_2$, a batch-level loss function for discriminatively boosted clustering optimization \cite{xie2016unsupervised}. The two loss functions work in tandem, enabling the joint refinement of image embeddings and cluster assignments in a self-paced and end-to-end manner.

\renewcommand{\algorithmicrequire}{\textbf{Input:}}
\renewcommand{\algorithmicensure}{\textbf{Output:}}
\begin{algorithm}[ht]
      \caption{Algorithm for Dual Advancement of Representations Learning and Clustering (\textit{DARLC}).}
      \label{pseudo-code}
      \begin{algorithmic}[1]
    \Require
      Images $\mathcal{X}\in \mathbb{R}^{N\times C\times H\times W}$; Seeding similarity matrix $\mathcal{S}\in \mathbb{R}^{N\times N}$; Denoise GAT $\mathcal{G}$; Maximum epochs $E_{max}$; Number of images \textit{N}; Number of clusters \textit{K}.
            \vspace{0.0pt}
    \renewcommand{\algorithmicrequire}{\textbf{Definition:}}
    \Require
    Parametes of $\cdot$ are denoted as $\Pi(\cdot)$; MIM branch $\mathcal{M}$; CL branch $\mathcal{C}$; Projection head $\mathcal{H}$; \textbf{UWL} (Uncertain Weight Loss function); \textbf{EMA} (Exponential Moving Average function).  
          \vspace{0.0pt}
    \Ensure
      Representations $Z\in \mathbb{R}^{N\times D}$;  Soft clustering $Q\in\mathbb{R}^{N\times K}$.
            \vspace{0.0pt}
    \State Compute smoothed image $\bar{\mathcal{X}}$ by Eq. (\ref{eq:data augmentation}). 
          \vspace{0.0pt}
    \While{$epoch < E_{max}$}
          \vspace{0.0pt}
        \For {$\mathbf{X}_b,\overline{\mathbf{X}}_b$ in $\mathcal{X},\hat{{\mathcal{X}}}$}  
              \vspace{0.0pt}
            \State Compute $\widehat{\mathbf{X}}_b$ by Eq. (\ref{eq:x_hat}) $\to \mathcal{L}_{rec}(\mathbf{X}_b, \widehat{\mathbf{X}}_b)$ by Eq. (\ref{eq:rec}).
                  \vspace{0.0pt}
            \State Compute $\mathbf{e}_b$, $\overline{\mathbf{e}}_b$ by Eq. (\ref{eq:e_i}) $\to \mathcal{L}_{clr}(\mathbf{e}_b,\overline{\mathbf{e}}_b)$ by Eq. (\ref{eq:clr}).
                  \vspace{0.0pt}
            \State $\mathcal{L}_{ssl}=\mathbf{UWL}(\mathcal{L}_{rec},\mathcal{L}_{clr})$. 
                  \vspace{0.0pt}
            \State Update $\Pi(\mathcal{M})$ using $\mathcal{L}_{ssl}$.
                  \vspace{0.0pt}
            \State Update $\Pi(\mathcal{C})$ with $\mathbf{EMA}(\Pi(\mathcal{M}))$.   
                  \vspace{0.0pt}
        \EndFor
              \vspace{0.0pt}
    \EndWhile
          \vspace{0.0pt}
    \While{(not converged) \& ($epoch < E_{max}$)} 
          \vspace{0.0pt}
        \State Compute $\widehat{\mathcal{X}}$ by Eq. (\ref{eq:x_hat}) $\to \mathcal{L}_{rec}(\mathcal{X},\widehat{\mathcal{X}})$ by Eq. (\ref{eq:rec}).
              \vspace{0.0pt}
        \State Compute $\mathbf{Z}$ by Eqs. (\ref{eq:e_i}), (\ref{eq:z_i}) $\to \mathcal{L}_{\ell ap}(\mathbf{Z}, \mathcal{S})$ by Eq. (\ref{eq:lap}).
              \vspace{0.0pt}
        \State SMM parameters inference using MAP-EM $\to \Theta$.
          \vspace{0.0pt}
        \State $Q=\text{SMM}(\mathbf{Z}|\Theta)$ $\to \mathcal{L}_{size}(Q), \mathcal{L}_{\ell \ell}(Q)$ by Eqs. (\ref{eq:size}), (\ref{eq:ll}).
          \vspace{0.0pt}
        \State $\mathcal{L}_1=\eta \mathcal{L}_{\ell ap}+(1-\eta)\mathbf{UWL}(-\mathcal{L}_{\ell \ell}, -\mathcal{L}_{size}, \mathcal{L}_{rec})$.
              \vspace{0.0pt}
        \State Update $\Pi(\mathcal{M}),\Pi(\mathcal{C}),\Pi(\mathcal{H})$ using $\mathcal{L}_1$.
              \vspace{0.0pt}
        \For {$\mathbf{X}_b,\overline{\mathbf{X}}_b$ in $\mathcal{X},\overline{{\mathcal{X}}}$}
              \vspace{0.0pt}
            \State Compute $P$ by Eq. (\ref{eq:target}) $\to \mathcal{L}_{kl}(P,Q)$ by Eq. (\ref{eq:kl}).
                   \vspace{0.0pt}
            \State Compute $\widehat{\mathbf{X}}_b$ by Eq. (\ref{eq:x_hat}) $\to \mathcal{L}_{rec}(\mathbf{X}_b, \widehat{\mathbf{X}}_b)$ by Eq. (\ref{eq:rec}).
                  \vspace{0.0pt}
            \State Compute $\mathbf{e}_b$, $\overline{\mathbf{e}}_b$ by Eq. (\ref{eq:e_i}) $\to \mathcal{L}_{clr}(\mathbf{e}_b,\overline{\mathbf{e}}_b)$ by Eq. (\ref{eq:clr}).
                  \vspace{0.0pt}
            \State $\mathcal{L}_2=\mathbf{UWL}(\mathcal{L}_{kl}, \mathcal{L}_{rec}, \mathcal{L}_{clr})$.
                  \vspace{0.0pt}
            \State Update $\Theta$ and $\Pi(\mathcal{M}),\Pi(\mathcal{C}),\Pi(\mathcal{H})$ using $\mathcal{L}_2$.
                  \vspace{0.0pt}
        \EndFor
              \vspace{0.0pt}
    \EndWhile \\
          \vspace{0.0pt}
    \Return $\mathbf{Z},Q$
      \end{algorithmic}
    \end{algorithm}
\subsection{Unified Self-supervised Representation Learning (Module I) }
\subsubsection{Denoising-based Data Augmentation.} We train a graph attention autoencoder \textbf{\(\mathcal{G}\)} to generate smoothed images, serving as augmented positive instances \cite{velivckovic2017graph,dong2022deciphering}. Initially, for each image, we construct an undirected and unweighted graph by treating pixels as nodes connected to their k-nearest neighbors. Specifically, for a given image \(\mathbf{X}\in \mathbb{R}^{C\times H \times W}\), where \(C\) is the number of channels, $H$ and $W$ are the height and width of the image, respectively. Let \(N_{pix}=H\times W\) denote the number of pixels, \(\mathbf{v}_{\iota}\in \mathbb{R}^C\) denote the pixel vector at location \(\iota\), \(\forall \iota \in \{1,2,...,N_{pix}\}\). The encoder in $\mathcal{G}$ comprises \(L\) layers. For each layer $t\in \{1,2,...L-1\}$, with the initial value $\mathbf{h}_{\iota}^{(0)}=\mathbf{v}_{\iota}$, the output  \(\mathbf{h}_{\iota}^{(t)}\in \mathbb{R}^{d_p}\) is calculated as follows: 
\begin{align}
\mathbf{h}_{\iota}^{(t)}=Leaky&ReLU(\sum_{\nu\in S_\iota}\mathbf{att}_{\iota \nu}(\mathbf{W}^{(t)}\mathbf{h}_\nu^{(t-1)})),  
\end{align}where \(\mathbf{W}^{(t)}\) represents the trainable weights of the \(t\)-th autoencoder layer, \(S_{\iota}\)  the set of node \(\iota\)'s neighbors within a pre-specified radius \(r\). The attention score, \(\mathbf{att}_{\iota\nu}\), between nodes \(\iota\) and \(\nu\) are computed as follows:
\begin{equation}
    \alpha_{\iota\nu}^{(t)}=\mathbf{w}_{att}^{(t)}LeakyReLU(\mathbf{W}^{(t)} [\mathbf{h}_\iota^{(t-1)})||\mathbf{h}_\nu^{(t-1)}]),
\end{equation}

\begin{equation}
    \mathbf{att}_{\iota\nu}^{(t)}=\frac{\mathrm{exp}(\alpha_{\iota\nu}^{(t)})}{\sum_{\iota\in S_\iota}\mathrm{exp}(\alpha_{\iota\nu}^{(t)})},
\end{equation}
where \(\mathbf{w}_{att}^{(t)}\) represents the trainable attention weights.  The decoder of $\mathcal{G}$ mirrors the encoder with tied weights. The total loss function is defined as:
\begin{equation}
   L_{denoise}=\sum_{\iota=1}^{N_{pix}}\Vert \mathbf{v}_\iota-\widetilde{\mathbf{h}}_\iota^{(0)} \Vert_2 ,
\end{equation}
where \(\widetilde{\mathbf{h}}_\iota^{(0)}\) denotes the reconstructed \(\mathbf{v}_\iota\) output by the decoder.  Once trained, \textbf{\(\mathcal{G}\)} is applied to any given image \(\mathbf{X}_i\), to generate a smoothed image $\overline{\mathbf{X}}_i$ for a given image \(\mathbf{X}_i\) as:
\begin{equation}
    \label{eq:data augmentation}
    \overline{\mathbf{X}}_i=\mathcal{G}(\mathbf{X}_i,\textbf{W},w_{att})
\end{equation}

\subsubsection{Unified Self-supervised Image Representation Learning.} This SSL model encompasses two branches: a MIM branch \(\mathcal{M}\) and a contrastive branch $\mathcal{C}$.  \(\mathcal{M}\) is an adapted version of MAE, specifically designed for generating image patch embeddings in the context of SNIs. In this adaption, the standard MAE encoder is replaced with a lightweight ViT encoder, denoted as $\mathcal{M}_E$, with four transformer blocks, four attention heads, and a higher masking ratio (80\%). Meanwhile, the original transformer-based MAE decoder, is substituted with a fully-connected linear decoder \(\mathcal{M}_{D}\). For any given image \(\mathbf{X}_i\), the regenrated image \(\widehat{\mathbf{X}}_i\)  is as follows:
\begin{equation}\label{eq:x_hat}
   \mathbf{\widehat{X}}_i=\mathcal{M}_D (\mathcal{M}_{E}(\mathbf{X}_i, \, \mathbf{W}_E), \mathbf{W}_D)
\end{equation}
The MIM branch loss for the current batch are: 
\begin{equation}\label{eq:rec}
    \mathcal{L}_{rec}=\frac{1}{N_b}\sum_{i=1}^{N_b}\frac{1}{N_{masked}}\sum_{j\in S_{masked}}(\mathbf{p}_{i,j}-\widehat{\mathbf{p}}_{i,j})^T(\mathbf{p}_{i,j}-\widehat{\mathbf{p}}_{i,j}),
\end{equation}
where $N_b$ is the batch size, \(\mathbf{W}_E\) and \(\mathbf{W}_D\) represent the parameters of the encoder and decoder, respectively.  \(N_{masked}\)  denotes the number of masked patches, \(S_{masked}\) the set of masked patches. $\mathbf{p}_{i,j}$ and $\widehat{\mathbf{p}}_{i,j}$ represent the original and regenerated \(j\)-th image patch of \(\mathbf{X}_i\), respectively. 

For the contrastive branch  $\mathcal{C}$, the \(N_b\) raw images form positive pairs with their respective smoothed images, and negative pairs with the other \(2N_b-2\)  images in the same batch. $\mathcal{C}$ is structured around a pseudo-siamese network with two encoders: an online encoder $\mathcal{C}_O$ and a target momentum encoder $\mathcal{C}_T$ . 
Both encoders share the identical network architecture as $\mathcal{P}_E$, with $\mathcal{C}_O$ and $\mathcal{P}_E$ having tied parameters. The parameters of $\mathcal{C}_T$ are updated using EMA. Concretely, let $\widetilde{\mathbf{W}}_E$ and $\overline{\mathbf{W}}_E$ denote the parameters of the $\mathcal{C}_O$ and $\mathcal{C}_T$, respectively. Then we have:
\begin{equation}
    \widetilde{\mathbf{W}}_E = \mathbf{W}_E,\,\overline{\mathbf{W}}_E=m\overline{\mathbf{W}}_E+(1-m)\widetilde{\mathbf{W}}_E
\end{equation}
Here, $m$ represents the momentum, fixed at 0.999. For each image $\mathbf{X}_i$ and its augmented counterpart $\overline{\mathbf{X}}_i$, their respective embedding vectors, $\mathbf{e}_i$ and $\overline{\mathbf{e}}_i \in \mathbb{R}^D$, are obtained as: $\mathbf{e}_i=\tilde{g}(\mathcal{C}_O(\mathbf{X}_i, \widetilde{\mathbf{W}}_E))$, $\overline{\mathbf{e}}_i=\bar{g}(\mathcal{C}_T(\overline{\mathbf{X}}_i,\overline{\mathbf{W}}_E)),$
\begin{align}\label{eq:e_i}
    \mathbf{e}_i=\tilde{g}(\mathcal{C}_O(\mathbf{X}_i, \widetilde{\mathbf{W}}_E)), \,\overline{\mathbf{e}}_i=\bar{g}(\mathcal{C}_T(\overline{\mathbf{X}}_i,\overline{\mathbf{W}}_E)),
\end{align}
where $\tilde{g}$ and $\bar{g}$ are linear mapping functions with trainable weights, and the weights of $\bar{g}$ are updated using EMA as well. The contrastive loss \(\mathcal{L}_{clr,i}\) is computed as :

\begin{align}\label{eq:clr}
    \mathcal{L}_{clr,i}= &-\log\frac{\mathbf{s}(\mathbf{e}_i,\overline{\mathbf{e}}_i)}{\sum_{k=1,k\neq i}^{N_b}\mathbf{s}(\mathbf{e}_i,\mathbf{e}_k)+\sum_{k=1}^{N_b}\mathbf{s}(\mathbf{e}_i,\overline{\mathbf{e}}_k)} \notag \\
    &-\log\frac{\mathbf{s}(\overline{\mathbf{e}}_i,\mathbf{e}_i)}{\sum_{k=1}^{N_b}\mathbf{s}(\overline{\mathbf{e}}_i,\mathbf{e}_k)+\sum_{k=1,k\neq i}^{N_b}\mathbf{s}(\overline{\mathbf{e}}_i,\overline{\mathbf{e}}_k)},
\end{align}
where $\mathbf{s}(\cdot ,\cdot)=\exp(\cos(\cdot,\cdot)/\tau)$, and \(\tau\) is a temperature coefficient, defaulting to 0.5. Consequently, the loss function \(\mathcal{L}_{clr}\) is defined as \(\mathcal{L}_{clr} = \frac{1}{N_b} \sum_{i=1}^{N_b} \mathcal{L}_{clr,i}\). Finally, $\mathcal{L}_{rec}$ and $\mathcal{L}_{clr}$ are dynamically integrated into the total SSL loss function $\mathcal{L}_{ssl}$ using the uncertain weights loss ($\mathbf{UWL}$) function~\cite{liebel2018auxiliary}:
\begin{align}
    \mathcal{L}_{ssl}&=\mathbf{UWL}(\mathcal{L}_{rec}, \mathcal{L}_{clr})\notag \\
    &=\frac1{2\sigma_1^2}\mathcal{L}_{rec}+\frac1{2\sigma_2^2}\mathcal{L}_{clr}
    +\log(1+\sigma_1^2)+\log(1+\sigma_2^2),
\end{align}
where \(\sigma_1 \)and \(\sigma_2\) are trainable noise parameters.

\subsection{Self-paced Deep Image Clustering (Module II)}

\subsubsection{Student’s t mixture model.} Let \(\mathbf{e}_i\) denote the \textit{Module I}-generated embedding vector for the $i$-th original image. We first map \(\mathbf{e}_i\) to $\mathbf{z}_i\in \mathbb{R}^D$ in a latent space wherein it is more conformed to a t-distribution. This mapping is achieved through a nonlinear projection head with batch normalization and scaled exponential linear unit (SELU) activation function:
\begin{equation}\label{eq:z_i}
    \mathbf{z}_i=SELU(BN(\mathbf{W}_p,\mathbf{e}_i)),
\end{equation}
    The set of these vectors, $\mathbf{Z}=\{\mathbf{z}_i\}_{i=1}^N\in \mathbb{R}^{N \times D}$ , is modeled using an SMM, whose components correspond to image clusters. Since extreme values are downweighed by SMM during parameter inference, this clustering is more robust to outliers and variances. The SMM is parameterized by \(\Theta=\{\theta_k{:}\pi_k,\, \mu_k,\,\Sigma_k,\, v_k,\, \forall k\in K\}\), where \(K\) represents the number of components and is assumed to be known or can be automatically inferred (see Section 4.1). Here, \(\pi_k,\, \mu_k,\, \Sigma_k,\, v_k\) denote the weight, mean, covariance matrix, and degree of freedom of the \(k\)-th component, respectively. The density function of \(\mathbf{z}_i\) is expressed as:
\begin{equation}
    p(\mathbf{z}_i|\Theta)=\sum_{k=1}^K\pi_k\operatorname{\mathbf{\phi}}(\mathbf{z}_i|\mu_k,\Sigma_k,v_k)
\end{equation}
For robust model inference, we use the maximum a posterior-expectation maximization (MAP-EM) algorithm. We apply a conjugate Dirichlet prior on \(\Pi=\{\pi_k, \forall k\in[1,K]\}\), denoted as \(\Pi \thicksim Dir(\Pi | \alpha^0)\), and a normal inverse Wishart (NIW) prior on \(\mu_k\) and \(\Sigma_k\), denoted as \(\mu_k, \Sigma_k \thicksim NIW(\mu_k, \Sigma_k | \mathrm{m}_0, \kappa_0, S_0, \rho_0)\) for all \(k \in [1, K]\).

    To simplify the inference, we rewrite the Student’s t density function $\phi$ as a Gaussian scale mixture by introducing an “artificial” hidden variable \(\zeta_{i,k}, \, \forall i\in [1, N]\ , \, \forall k\in [1, \, K]\) that follows a Gamma distribution parameterized by \(v_k\):
\begin{align}
    \phi(\mathbf{z}_i|\mu_k,&\Sigma_k,v_k)= \int\mathcal{N}\left(\mathbf{z}_i\bigg|\mu_k,\frac{\Sigma_k}{\zeta_{i,k}}\bigg)\Gamma\left(\zeta_{i,k}\bigg|\frac{v_k}{2},\frac{v_k}{2}\right)d\zeta_{i,k}\right.
\end{align}
    We further introduce a hidden variable \(\xi_i\) to represent the component membership of \(\mathbf{z}_i\). The posterior complete data log likelihood is then expressed as:
\begin{align}
    &\ell_c(\Theta)=\log P(\mathbf{Z}, \zeta, \xi |\Theta)\notag \\
    &=\mathrm{logDir}(\Pi|\alpha^0)+\sum_klogNIW(\mu_k, \Sigma_k|m_0, \kappa_0, S_0, \rho_0)\notag+ \\&\sum_i\sum_k[
    II(\xi_i=k)(\log\pi_k+\log\Phi(\mathbf{z}_i, \zeta_{i, k}|\mu_k, \Sigma_k, v_k))]
\end{align}
    In the $t$-th iteration of the E-step, the expected sufficient statistics $\overline{\xi_{i,k}}^{(t)}$ and $\overline{\zeta_{i,k}}^{(t)}$ are derived based on $\Theta^{(t-1)}$. In the subsequent M-step, $\Theta^{(t-1)}$ is updated to $\Theta^{(t)}$ by maximizing the auxiliary function $Q(\Theta,\, \Theta^{(t-1)})=E(\ell_c(\Theta)|\Theta^{(t-1)})$. These two steps are alternated until either convergence is reached or a predefined maximum number of iterations is attained. Refer to Supplementary 1.1 for details of the model inference.

\subsubsection{Self-paced Joint Optimization of Image Embeddings and Cluster Assignments.}
Two loss functions, $\mathcal{L}_1$ and $\mathcal{L}_2$, are calculated based on clustering results for updating parameters of both \textit{Module I}\textbf{ }and the SMM through loss gradient backpropagation. This iterative process progressively improves the clustering-oriented image embeddings and clustering results. Upon completing the inference of SMM parameters $\widetilde{\Theta}$ in each epoch, an epoch-level loss $\mathcal{L}_1$ is calculated for updating parameters of \textit{Module I}:
\begin{equation}
    \mathcal{L}_1=\eta\mathcal{L}_{\ell ap}+(1-\eta)\mathbf{UWL}(-\mathcal{L}_{\ell \ell},-\mathcal{L}_{size},\mathcal{L}_{rec}).
\end{equation}
    Here, \(\mathcal{L}_{\ell ap}\) is a Laplacian regularization term that promotes the similarities among image embeddings \(\mathbf{Z}\) to be consistent with a seeding image-image similarity matrix \(\mathcal{S}\), informing the initial training phase. The derivation of \(\mathcal{S}\) is detailed in Supplementary 1.2. \(\mathcal{L}_{\ell ap}\) is defined as follows:
\begin{align}\label{eq:lap}
    \mathcal{L}_{\ell ap}&=Tr\left(Z^{T}\left(I-\mathcal{D}^{-\frac{1}{2}}\mathcal{S}\mathcal{D}^{-\frac{1}{2}}\right)Z\right),
\end{align}
where \(\mathcal{D}\) is the degree matrix of \(\mathcal{S}\), and $\eta$, initially set at 0.5, decays over the training course so that the influence of \(\mathcal{S}\) is gradually reduced. \(\mathcal{L}_{\ell \ell}\) represents the log likelihood of the embeddings given the estimated SMM parameters $\widetilde{\Theta}$:
\begin{align}\label{eq:ll}
    {\mathcal L}_{\ell\ell}=\sum_{\mathrm{i}=1}^{\mathrm{N}}\mathrm{log}\left[\sum_{k}q_{i,k}\right],
\end{align}

\begin{table*}[!htbp]
\centering
\caption{Clustering performance of \textit{DARLC} and benchmark methods across 12 datasets. `ST' represents spatial transcriptome-based gene expression images, and `MSI' represents mass spectrometer-based metabolite concentration images, quantified using DBIE and DBIP scores. `*' represents sparsified real-world images with 90\% random pixel masking, quantified using NMI and ARI scores. Lower DBIE (DBIP) and higher NMI (ARI) scores indicate better performance. The best score for each dataset is \textbf{bolded}, and the second-best score is \underline{underlined}. The score standard deviation is subscripted.}
\label{tab:comparison}

\renewcommand{\arraystretch}{1.45}
\fontsize{8.95pt}{8.95 pt}\selectfont
\setlength{\tabcolsep}{1.7 pt}
\begin{tabular}{|l|cccccc|cccc|cc|}
\hline
\rowcolor{mygray}
& \multicolumn{10}{c|}{DBIE$\downarrow$}
& \multicolumn{2}{c|}{NMI (\%)$\uparrow$}\\
\cline{2-13} 
\rowcolor{mygray}
\multirow{-2}{*}{Method} & \multicolumn{6}{c|}{ST-hDLPFC-\{1-6\}}& ST-hMTG & ST-hBC & ST-mEmb & MSI-mBrain & STL-10* & CIFAR-10* \\


\hline
\hline
\multirow{1}{*}{DEC} & $\text{12.32}_{0.97}$ & $\text{11.05}_{0.59}$ & $\text{11.58}_{1.20}$ & $\text{11.22}_{1.08}$ & $\text{11.44}_{0.70}$ & $\text{11.42}_{0.65}$ & $\underline{\text{9.42}}_{0.10}$ & $\text{9.41}_{0.77}$ & $\underline{\text{8.05}}_{\text{0.25}}$ & $\text{5.30}_{0.13}$ & $\text{13.85}_{0.19}$ & $\text{6.89}_{0.14}$ \\

\multirow{1}{*}{DAGMM} & $\text{46.71}_{42.62}$ & $\text{54.80}_{42.75}$ & $\text{26.07}_{5.72}$ & $\text{39.47}_{12.57}$ & $\text{20.64}_{17.48}$ & $\text{42.83}_{19.63}$ & $\text{38.27}_{18.09}$ & $\text{74.04}_{42.59}$ & $\text{19.29}_{10.78}$ & $\text{6.51}_{0.47}$ & $\text{4.13}_{0.26}$ & $\text{1.36}_{0.03}$ \\

\multirow{1}{*}{EDESC} & $\text{11.94}_{1.08}$ & $\text{13.53}_{1.53}$ & $\underline{\text{11.18}}_{0.99}$ & $\text{10.17}_{1.70}$ & $\underline{\text{9.94}}_{1.44}$ & $\underline{\text{10.04}}_{1.12}$ & $\text{10.06}_{0.32}$ & $\text{9.94}_{0.40}$ & $\text{8.24}_{0.18}$ & $\text{4.91}_{0.58}$ & $\text{11.02}_{0.48}$ & $\text{2.84}_{0.12}$ \\

\multirow{1}{*}{IDCEC} & $\underline{\text{10.51}}_{0.34}$ & $\text{10.83}_{0.49}$ & $\text{11.23}_{0.50}$ & $\text{10.60}_{0.34}$ & $\text{12.30}_{0.94}$ & $\text{12.95}_{1.10}$ & $\text{9.58}_{0.03}$ & $\underline{\text{9.34}}_{0.09}$ & $\text{8.44}_{0.14}$ & $\underline{\text{4.44}}_{0.14}$ & $\underline{\text{14.90}}_{0.80}$ & $\text{3.05}_{0.08}$ \\

\multirow{1}{*}{CC} & $\text{17.05}_{0.03}$ & $\text{19.05}_{0.05}$ & $\text{18.77}_{0.04}$ & $\text{19.53}_{0.05}$ & $\text{20.26}_{0.02}$ & $\text{17.41}_{0.03}$ & $\text{22.61}_{0.33}$ & $\text{36.66}_{2.21}$ & $\text{26.32}_{3.34}$ & $\text{6.72}_{0.34}$ & $\text{10.24}_{0.28}$ & $\text{9.96}_{0.51}$ \\

\multirow{1}{*}{DCP} & $\text{11.61}_{0.11}$ & $\text{12.28}_{0.39}$ & $\text{11.92}_{0.27}$ & $\text{11.33}_{0.86}$ & $\text{12.09}_{0.66}$ & $\text{11.27}_{0.03}$ & $\text{10.07}_{0.01}$ & $\text{9.87}_{0.01}$ & $\text{8.60}_{0.02}$ & $\text{5.95}_{0.87}$ & $\text{10.81}_{1.41}$ & $\text{2.21}_{0.19}$  \\

\multirow{1}{*}{CVCL} & $\text{39.52}_{11.33}$ & $\text{31.00}_{0.51}$ & $\text{30.77}_{1.70}$ & $\text{31.15}_{3.20}$ & $\text{39.85}_{7.30}$ & $\text{30.96}_{2.35}$ & $\text{17.93}_{0.77}$ & $\text{20.87}_{0.17}$ & $\text{16.87}_{0.29}$ & $\text{6.16}_{0.17}$ & $\text{9.72}_{0.22}$ & $\text{6.07}_{0.22}$ \\

\multirow{1}{*}{iBOT-C} & $\text{12.69}_{0.04}$ & $\underline{\text{10.76}}_{0.05}$ & $\text{12.81}_{0.02}$ & $\underline{\text{10.11}}_{0.01}$ & $\text{10.71}_{0.03}$ & $\text{11.31}_{0.01}$ & $\text{25.86}_{0.14}$ & $\text{41.24}_{4.35}$ & $\text{30.23}_{4.92}$ & $\text{7.65}_{0.35}$ & $\text{10.76}_{0.34}$ & $\underline{\text{10.74}}_{0.14}$ \\
\hline
\multirow{1}{*}{\textit{DARLC}} & \textbf{$\text{7.65}_{0.40}$} & \textbf{$\text{8.06}_{0.37}$} & \textbf{$\text{7.90}_{0.37}$} & \textbf{$\text{7.78}_{0.33}$} & \textbf{$\text{7.41}_{0.36}$} & \textbf{$\text{8.09}_{0.55}$} & \textbf{$\text{9.33}_{0.04}$} & $\textbf{\text{9.17}}_{0.04}$ & \textbf{$\text{7.70}_{0.54}$} & \textbf{$\text{4.30}_{0.11}$} & \textbf{$\text{15.26}_{0.29}$} & \textbf{$\text{10.99}_{1.36}$} \\
\hline\hline

\rowcolor{mygray}
& \multicolumn{10}{c|}{DBIP$\downarrow$}
& \multicolumn{2}{c|}{ARI (\%)$\uparrow$}\\
\cline{2-13} 
\rowcolor{mygray}
\multirow{-2}{*}{Method} & \multicolumn{6}{c|}{ST-hDLPFC-\{1-6\}}& ST-hMTG & ST-hBC & ST-mEmb & MSI-mBrain & STL-10* & CIFAR-10* \\

\hline
\hline

DEC & $\text{3.04}_{0.24}$ & $\underline{\text{2.52}}_{0.08}$ & $\text{3.71}_{0.53}$ & $\text{3.32}_{0.19}$ & $\text{3.52}_{0.59}$ & $\underline{\text{2.42}}_{0.08}$ & $\text{4.13}_{0.04}$ & $\text{7.76}_{1.72}$ & $\text{6.68}_{0.37}$ & $\text{12.52}_{1.39}$ & $\underline{\text{6.89}}_{0.14}$ & $\text{2.33}_{0.11}$ \\
DAGMM & $\text{31.03}_{18.08}$ & $\text{22.75}_{12.10}$ & $\text{27.17}_{13.81}$ & $\text{20.06}_{2.27}$ & $\text{16.14}_{13.59}$ & $\text{20.14}_{9.84}$ & $\text{52.18}_{7.01}$ & $\text{136.40}_{12.43}$ & $\text{195.54}_{46.48}$ & $\text{12.27}_{1.09}$ & $\text{1.63}_{0.07}$ & $\text{0.87}_{0.06}$ \\
EDESC & $\text{2.70}_{0.36}$ & $\text{2.66}_{0.38}$ & $\underline{\text{2.85}}_{0.40}$ & $\text{3.11}_{0.14}$ & $\underline{\text{2.75}}_{0.30}$ & $\text{2.77}_{0.30}$ & $\text{4.22}_{0.09}$ & $\text{6.05}_{1.04}$ & $\text{6.10}_{0.15}$ & $\text{13.13}_{1.62}$ & $\text{4.63}_{1.76}$ & $\text{1.53}_{0.01}$ \\
IDCEC & $\text{2.76}_{0.26}$ & $\text{2.69}_{0.32}$ & $\text{3.13}_{0.23}$ & $\text{2.99}_{0.20}$ & $\text{2.98}_{0.26}$ & $\text{3.07}_{0.19}$ & $\text{3.88}_{0.09}$ & $\text{5.06}_{0.21}$ & $\text{7.47}_{0.38}$ & $\text{12.75}_{2.24}$ & $\text{6.61}_{0.06}$ & $\text{1.61}_{0.04}$ \\
CC & $\text{4.12}_{0.02}$ & $\text{4.35}_{0.01}$ & $\text{3.18}_{0.00}$ & $\text{3.58}_{0.00}$ & $\text{2.93}_{0.01}$ & $\text{3.96}_{0.04}$ & $\text{6.14}_{0.02}$ & $\text{13.17}_{1.81}$ & $\text{9.03}_{0.22}$ & $\text{11.55}_{0.40}$ & $\text{4.13}_{0.15}$ & $\text{3.97}_{0.32}$ \\
DCP & $\underline{\text{2.59}}_{0.06}$ & $\text{2.55}_{0.06}$ & $\text{3.37}_{0.03}$ & $\underline{\text{2.96}}_{0.16}$ & $\text{3.00}_{0.13}$ & $\text{2.87}_{0.16}$ & $\underline{\text{3.33}}_{0.02}$ & $\underline{\text{5.03}}_{0.01}$ & $\underline{\text{5.88}}_{\text{0.01}}$ & $\text{11.91}_{0.63}$ & $\text{3.04}_{0.34}$ & $\text{0.68}_{0.18}$ \\
CVCL & $\text{8.92}_{2.02}$ & $\text{6.69}_{0.81}$ & $\text{7.13}_{0.84}$ & $\text{5.45}_{0.63}$ & $\text{6.32}_{1.07}$ & $\text{4.97}_{0.02}$ & $\text{4.17}_{0.27}$ & $\text{5.55}_{0.20}$ & $\text{6.61}_{0.26}$ & $\text{12.41}_{0.47}$ & $\text{4.56}_{0.01}$ & $\text{1.95}_{0.01}$ \\
iBOT-C & $\text{3.48}_{0.03}$ & $\text{2.96}_{0.02}$ & $\text{4.87}_{0.01}$ & $\text{3.38}_{0.01}$ & $\text{3.57}_{0.03}$ & $\text{3.47}_{0.02}$ & $\text{7.83}_{0.02}$ & $\text{15.41}_{2.31}$ & $\text{11.26}_{0.13}$ & $\underline{\text{11.36}}_{0.33}$ & $\text{4.56}_{0.29}$ & $\underline{\text{4.56}}_{0.10}$ \\
\hline
\textit{DARLC} & \textbf{$\text{2.24}_{0.06}$} & \textbf{$\text{2.19}_{0.03}$} & \textbf{$\text{2.23}_{0.00}$} & \textbf{$\text{2.14}_{0.08}$} & \textbf{$\text{2.45}_{0.11}$} & \textbf{$\text{2.11}_{0.20}$} & \textbf{$\text{2.93}_{0.02}$} & \textbf{$\text{4.07}_{0.04}$} & \textbf{$\text{3.15}_{0.33}$} & \textbf{$\text{10.56}_{1.03}$} & \textbf{$\text{7.13}_{0.44}$} & \textbf{$\text{5.08}_{0.24}$} \\
\hline
\end{tabular}

\end{table*}

\begin{equation}\label{prob}
    q_{i, k} = \pi_{k}\phi(\mathbf{z}_{i}|\mu_{k},\Sigma_{k},v_{k}), \forall i\in[1,\, N], \, \forall k\in [1,\, K].
\end{equation}
    \(\mathcal{L}_{size}\) penalizes empty and tiny clusters, while exempting those whose size exceeds a predefined threshold \(\upsilon\) so that image assignments is not overly uniform:
\begin{equation}\label{eq:size}
    \mathcal{L}_{size}=\sum_{k=1}^{K}-J_{k}\mathrm{log}J_{k},J_k=\begin{cases}\frac{\Sigma_i^Nq_{i,k}}{N}&,ifJ_k\leq\upsilon\\1&,otherwise\end{cases}
\end{equation}
$\mathcal{L}_{rec}$, defined in Equation \ref{eq:rec}, aims to enhance the local-context awareness of embeddings. Subsequently, within the same epoch, a batch-level loss \(\mathcal{L}_2=\mathbf{UWL}(\mathcal{L}_{kl}, \mathcal{L}_{rec},\mathcal{L}_{clr})\) is utilized to update \textit{Module I} and SMM parameters across successive 
 batches. Here, $\mathcal{L}_{rec}$ and $\mathcal{L}_{clr}$ remains same as in Equations \ref{eq:rec} and \ref{eq:clr} except being calculated on the batch-level. \(\mathcal{L}_{kl}\)  boosts high-confidence images, incrementally grouping similar instances while separating dissimilar ones:
\begin{equation}\label{eq:kl}
\mathcal{L}_{kl}=KL(\mathcal{P}|Q)=\sum_{i}^{N}\sum_{j}^{K}\vmathbb{p}_{i,j}\log\frac{\vmathbb{p}_{i,j}}{\vmathbb{q}_{i,j}}, 
\end{equation}
    \begin{equation}\label{eq:target}
    \text{where}\ \vmathbb{q}_{i,k}=\frac{q_{i,k}}{\sum_cq_{i,c}},\vmathbb{p}_{i,k}=\frac{\vmathbb{q}_{i,k}^{2}/\sum_{i}\vmathbb{q}_{i,k}}{\sum_{c}\left(\vmathbb{q}_{i,c}^{2}/\sum_{i}\vmathbb{q}_{i,c}\right)}
\end{equation}
    Here, $q_{i,k}$ is same as in Equation \ref{prob}, $\vmathbb{q}_{i,k}$ represents the probability of assigning $i$-th image to the $k$-th SMM component, and $\vmathbb{p}_{i,k}$ an auxiliary target distribution that boosts up high-confidence images. After this joint optimization, the training progresses to the next epoch, iterating until the end of the training process. The mathematical derivations of gradients of $\mathcal{L}_1$ and $\mathcal{L}_2$ with respect to $\mathbf{W}_E$, $\mathbf{W}_D$ and $\Theta$ are detailed in Supplementary 1.3.

\section{Experiments}
\begin{figure*}[h]
  \centering
  \includegraphics[width=\linewidth]{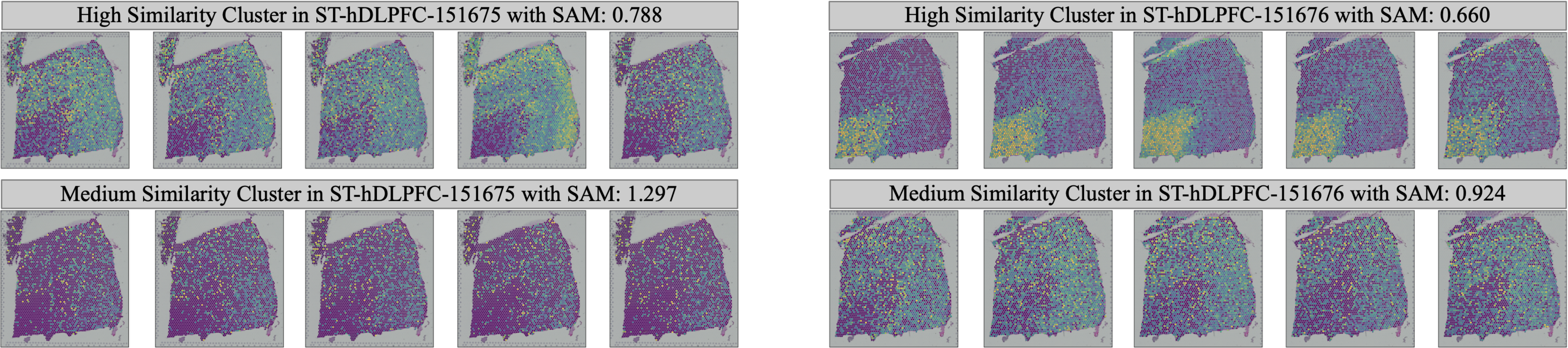}
  \caption{SGEPs from clusters generated by \textit{DARLC} in dataset (ST-hLDPFC-\{5-6\}) with high and medium intra-cluster similarity.}
  \label{sam}
\end{figure*}
\subsection{Experimental Settings}
\begin{table}[t]
    \centering
    \caption{Gene-gene interaction prediction results for three gene image embeddings. The superior method is \textbf{bolded}.}
    \fontsize{9.pt}{9.pt}\selectfont
    \renewcommand{\arraystretch}{1.4}  
    \setlength{\tabcolsep}{5.2pt}  
    \begin{tabularx}{\linewidth}{|l|cccccc|}
        \hline
        \rowcolor{mygray}
        & \multicolumn{6}{c|}{ACC (\%)} \\ 
        \cline{2-7} 
        \rowcolor{mygray}
        \multirow{-2}{*}{Case} & \multicolumn{6}{c|}{ST-hDLPFC-\{1-6\}} \\ \hline
        \hline
        iBOT & 66.58& 68.73& 68.53& 65.88& 68.03& 67.48\\
        \textit{DARLC}-C1 & 76.18& 77.20 &74.71 & 73.64& 77.85& 71.50\\
        \hline
        \textit{DARLC}-full& \textbf{78.11}& \textbf{78.97}& \textbf{77.66}& \textbf{78.74}& \textbf{78.65}& \textbf{73.59}\\ 
         \hline
    \end{tabularx}
    
    \label{gene interaction}
\end{table}
\subsubsection{Datasets.} 
To comprehensively analyze DARLC, we utilize 9 real ST datasets from 6 human dorsolateral prefrontal cortex  slices (ST-hDLPFC-\{1-6\}), 1 human middle temporal gyrus slice (ST-hMTG), 1 human breast cancer slice (ST-hBC), and 1 mouse embryo slice (ST-mEmb). These ST datasets display the spatial expression levels of the entire genome across the corresponding tissue, approximately 18,000 75x75 SGEP images per dataset. Additionally, we use 1 real mass spectrometry imaging dataset from a mouse brain (MSI-hBrain). This dataset displays the spatial expression levels of the entire metabolome, containing 847 110x59 SNIs. We also utilize two artificially sparsified STL-10* and CIFAR-10* datasets with 90\% pixels masked, which comprise 13,000 96x96x3 and 60,000 32x32x3 images, respectively. All datasets are available and detailed description can be found in Supplementary 1.4.

\subsubsection{Data quality control and preprocessing.} We conform to the conventional procedure for preprocessing ST data, as implemented in the SCANPY package \cite{wolf2018scanpy}. Specifically, we first remove mitochondrial and External RNA Controls Consortium spike-in genes. Then, genes detected in fewer than 10 spots are excluded. To preserve the spatial data integrity, we do not perform quality control on spatial spots. Finally, the gene expression counts are normalized by library size, followed by log-transformation. As for MSI dataset, we utilize traditional precedure total ion current normalization \cite{fonville2012robust}.

\subsubsection{Cluster Number Inference.} Given the number of image clusters is not known a priori, \textit{DARLC} can estimate this number using a seeding similarity matrix $\mathcal{S}\in\mathbb{R}^{N \times N}$, where $N$ represents the number of images \cite{Clunum}. $\mathcal{S}$ is transformed to a graph Laplacian matrix (refer to Supplementary 1.2), $L^\prime\in\mathbb{R}^{N\times N}$ as follows:
\begin{equation}
    L^\prime=L+L^2,
\end{equation}
where $L=\mathbb{D}-\mathcal{S}$, $\mathbb{D}$ is $\mathcal{S}$'s degree matrix, and $L^2$ aims to enhance the similarity structure. Then the normalized Laplacian matrix can be obtained as:
\begin{equation}
    \mathbb{L}=\mathbb{D}^{-\frac{1}{2}}L^\prime\mathbb{D}^{-\frac{1}{2}}.
\end{equation}
The eigenvalues of $\mathbb{L}$ are then ranked as $\lambda_{(1)} \le \lambda_{(2)} \le \cdots \le \lambda_{(N)}$. The number of clusters, denoted as $K$, is inferred as:
\begin{equation}
    K={\rm argmax}_i\left\{\lambda_{\left(i\right)}-\lambda_{\left(i-1\right)}\right\},\ \ \ \ i=2,\ 3,\ \cdots,\ N.
\end{equation}

\subsubsection{Baselines.} The image clustering benchmark methods include two classic deep clustering method, DEC \cite{xie2016unsupervised} and DAGMM \cite{zong2018deep}, as well as five SOTA methods that either improve DEC (e.g.,EDESC \cite{cai2022efficient} and IDCEC~\cite{lu2022improved}) or incorporate CL (e.g., CC \cite{li2021contrastive}, DCP \cite{lin2022dual}, and CVCL \cite{chen2023deep}). In addition, we include a benchmark method consisting of iBOT \cite{zhou2021ibot}, which integrates CL and MIM for representation learning, and a boosted GMM for clustering, denoted as iBOT-C.

\subsubsection{Implementation Details.} 
The GAT for data augmentation adopts an encoder including a single attention head with C-512-30 network structure, and a symmetric decoder. In \textit{Module I}, the shared encoder structure is a ViT comprising four transformer blocks, each having four attention heads, for processing $75\times 75$ input images segmented into $4\times 4$ patches in our case. The MIM decoder follows a D-128-256-512-1024-75*75 residual network. The \textit{Module I} is pretrained for 50 epochs with a learning rate of 0.001. The nonlinear projection head that bridges \textit{Module I} and \textit{II }is a two-layer MLP for normalizing image representations to a dimension size of 32. The iterative joint optimization of representation learning and clustering continues for 50 epochs using Adam optimizer. Given the absence of ground truth in gene cluster labels, we heuristically determine the number of gene clusters for our experiments to achieve an average cluster size of 30 genes, approximating the typical size of a gene pathway ~\cite{zhou2012intpath}.

\subsubsection{Evaluation Metrics.}
Without ground truth cluster labels, we evaluate the clustering results using the Davies-Bouldin index (DBI) metric ~\cite{davies1979cluster}:
\begin{align}
    \begin{split}
\mathrm{DBI}=\mathrm{~}\frac1K\sum_{i=1}^K\max_{i\neq j}\frac{d_i+d_j}{d_{(i,j)}},
    d_i=\frac1{|C_i|}\sum_{j=1}^{|C_i|}\delta_{c_{i},j},
    \end{split}
\end{align}
where $K$ is the number of clusters, $C_i$ the samples in cluster $i$, $\delta_{i,j}$ the distance between instances $i$ and $j$, $c_i$ the centroid of cluster $i$. Cluster width $d_i$ is the mean intra-cluster distance to $c_i$, and $d_{(i,j)}=\delta_{c_i,c_j}$ measures the distance between clusters $i$ and $j$. DBI quantifies the clustering efficiency by measuring the ratio of intra-cluster compactness to inter-cluster separation, with lower scores indicating better clustering. To evaluate clustering from different perspectives, we use two DBI metrics, DBIP and DBIE, based on Pearson and Euclidean distances, respectively~\cite{song2022detecting}.

\begin{table}[t]
    \centering
    \caption{Spatial cofunctional gene clustering results using three image embeddings, with the best method in \textbf{bolded}}.
    \fontsize{9.pt}{9.pt}\selectfont
    \renewcommand{\arraystretch}{1.4}
    \setlength{\tabcolsep}{5.2pt}
    \begin{tabularx}{\linewidth}{|l|cccccc|}
        \hline
        \rowcolor{mygray}
        & \multicolumn{6}{c|}{NMI (\%)$\uparrow$} \\ 
        \cline{2-7} 
        \rowcolor{mygray}
        \multirow{-2}{*}{Case} & \multicolumn{6}{c|}{ST-hDLPFC-\{1-6\}} \\ \hline
        \hline
        iBOT &20.56 & 15.98& 15.41& 27.05& 23.17& 25.62\\
        \hline
        \textit{DARLC}-C1 & 66.81& 67.81& 70.91& 65.13& 74.62& 75.85\\
        \textit{DARLC}-full& \textbf{78.11}& \textbf{71.28}& \textbf{79.23}& \textbf{74.34}& \textbf{75.20}& \textbf{75.91}\\ \hline
        
    \end{tabularx}

    \label{embedding analysis}
\end{table}

\subsection{Clustering Sparse, Noisy Images of Spatial Gene Expressions}
Table \ref{tab:comparison} showcase the performance of \textit{DARLC}, compared to eight benchmark methods, in clustering SNIs across 12 datasets, evaluated by DBIE and DBIP for unlabeled, NMI and ARI for labeled datasets, the experiment is repeated ten times to obtain the mean and standard deviation of each method's scores. \textit{DARLC} consistently scores lowest in DBIE and DBIP for all unlabeled datasets and highest in NMI and ARI for labeled datasets, highlighting its superiority in generating clusters consisting of spatially similar and coherent images. This superiority can be attributed to \textit{DARLC}'s features in integrating MIM and CL, generating more plausible augmented data, and the robust and adaptable clustering algorithm, as substantiated in our ablation study. In contrast, benchmark methods adopt varied strategies: IDCEC and EDESC leverage a convolutional autoencoder for extracting visual features; iBOT+boosted GMM adopts conventional data augmentation; CC, CVCL and DCP rely solely on CL for representation learning; DAGMM employs an outlier-sensitive GMM for clustering. However, these methods generally demonstrate unstable and suboptimal performance. Overall, compared to the best-performing benchmark method, \textit{DARLC} achieves an average reduction of approximately 15.98\% in DBIE and 18.44\% in DBIP across all ST and MSI datasets. Finally, \textit{DARLC} clusters are divided into high and medium-quality groups using spectral angle mapper (SAM) metric scores~\cite{kruse1993spectral}, with lower SAM indicating greater intra-cluster similarity. To provide a visual illustration of \textit{DARLC}'s clustering performance, we randomly select one cluster from both the high- and medium-quality groups within each of the ST-hDLPFC-\{5-6\} datasets. From each of the selected clusters, we then randomly select five genes to be displayed in Figure \ref{sam}. Figure \ref{sam} clearly demonstrates that \textit{DARLC} effectively groups images with similar expression patterns into the same cluster.

\subsection{Evaluating \textit{DARLC}-generated Gene Image Representations}
In this section, we present a comprehensive evaluation of gene image representations produced by the fully implemented \textit{DARLC} model (denoted as \textit{DARLC}-full). First, we assess whether representations generated by \textit{DARLC}-full capture corresponding biosemantics, particularly through pathway enrichment analysis, compared to original gene expressions. Furthermore, we extend our investigation to specific critical downstream tasks: predicting interactions between genes and clustering genes based on spatial cofunctionality. These evaluations are conducted across six distinct datasets (ST-hDLPFC-\{1-6\}). Additionally, gene image embeddings generated by iBOT and a variant of \textit{DARLC} (denoted as \textit{DARLC}-C1), which is deprived of \textit{Module II}, serve as baselines.

\subsubsection{Gene-gene Interaction Prediction.} We employ an MLP-based classifier for predicting gene pair interactions. This is achieved by linear probing using gene image representations generated by \textit{DARLC} and baseline methods (see Supplementary 1.5). We follow the methodology in \cite{du2019gene2vec}, which is based on the Gene Ontology, to acquire the gene-gene interaction ground truth. Theoretically, gene image representations with richer semantic meanings should yield more accurate predictions. As shown by the accuracy scores in Table \ref{gene interaction},  the classifier yields the most accurate predictions (77.62\%±1.85\%) using embeddings generated by \textit{DARLC}-full, and the second most accurate predictions (75.18\%±2.17\%) using embeddings generated by \textit{DARLC}-C1, followed by the predictions using iBOT-generated embeddings (67.54\%±1.03\%).

\subsubsection{Spatially Cofunctional Gene Clustering.} Spatially cofunctional genes are those belonging to the same gene family and exhibit similar spatial expression patterns ~\cite{demuth2006evolution}. Their family identities can serve as labels to evaluate the quality of gene image embeddings via clustering. Specifically, our evaluation involves five spatially cofunctional genes from each of the HLA, GABR, RPL, and MT gene families (see Supplementary Figure 1).  The Leiden algorithm \cite{traag2019louvain} is used to cluster gene image embeddings generated by \textit{DARLC}-full and the baseline methods.  The clustering results, evaluated using the normalized mutual information (NMI) and adjusted rand index (ARI) scores, as shown in Table \ref{embedding analysis} and Supplementary Table 3, demonstrate that Leiden yields the most accurate clustering with \textit{DARLC}-full and the second most accurate with \textit{DARLC}-C1.

In summary, these results collectively highlight the effectiveness of the joint clustering (i.e., \textit{DARLC}-full surpasses \textit{DARLC}-C1) and GAT-based data augmentation (i.e., \textit{DARLC}-C1 surpasses iBOT) in enhancing the quality of gene image representations. 

\subsection{Ablation Study}
Here, we conduct a series of ablation studies on six ST datasets (ST-hDLPFC-\{1-6\}) to investigate the contributions of \textit{DARLC}'s components in image clustering. The results, detailed in DBIE and DBIP scores, are presented in Table \ref{ablation study} and Supplementary Table 4. Notably, \textit{DARLC}'s performance declines most with the CL branch removal (``w/o CLR''), followed by the elimination of \textit{Module II} (``w/o SMM") and the substitution of the robust SMM with an outlier-sensitive GMM (``SMM $\rightarrow$ GMM"). We also observed that employing traditional image smoothing methods such as Gaussian Kernel Smoothing (GKS) instead of GAT (``GAT $\rightarrow$ GKS'') on SNIs leads to a decline in \textit{DARLC}'s performance. Additionally, removing either $\mathcal{L}_{\ell ap}$ (``w/o $\mathcal{L}_{\ell ap}$") or $\mathcal{L}_{size}$ (``w/o $\mathcal{L}_{size}$") from the optimization decreases \textit{DARLC}'s performance, indicated by higher DBIE and DBIP scores. Lastly, ``w/o Pretraining" showcases \textit{DARLC}'s performance at the same clustering training epoch as the complete model but without the initial ``warm up" pretraining of \textit{Module I}. The relative underperformance in the ``w/o Pretraining" scenario suggests a slower training convergence compared to the complete model.

\begin{table}[t]
    \centering
    \caption{Ablation study results across six datasets for components in \textit{Module I \&II} and regularization terms. The best result is \textbf{bolded}.}
    \fontsize{9.5pt}{9.5pt}\selectfont
    \renewcommand{\arraystretch}{1.4}  
    \setlength{\tabcolsep}{4.3pt}
    \begin{tabularx}{\linewidth}{|l|cccccc|}
        \hline
        \rowcolor{mygray}
        & \multicolumn{6}{c|}{DBIE$\downarrow$} \\ 
        \cline{2-7}
        \rowcolor{mygray}
        \multirow{-2}{*}{Case}
        & \multicolumn{6}{c|}{ST-hDLPFC-\{1-6\}} \\ 
        \hline
        \hline
        
        w/o CLR &14.35 &16.99 &8.70 &9.23 &10.36 &15.12 \\
        GAT $\rightarrow$ GKS &9.35 &10.37 &8.98 &9.08 &8.90 &8.95 \\
        w/o SMM &11.84 &12.19 &11.33 &11.81 &11.49 &12.35 \\
        SMM $\rightarrow$ GMM &11.49 &11.36 &11.16 &11.31 &11.05 &11.37 \\
        w/o $\mathcal{L}_{\ell ap}$ &8.32 &10.76 &9.02 &8.10 &9.22 &8.12 \\
        w/o $\mathcal{L}_{size}$ &10.38 &8.60 &8.11 &9.34 &8.00 &8.28 \\
        w/o Pretraining &8.22 &8.74 &8.38 &8.18 &7.87 &9.09 \\\hline
        \textit{DARLC}-full &\textbf{7.65} &\textbf{8.06} &\textbf{7.90} &\textbf{7.78} &\textbf{7.41} &\textbf{8.09} \\ \hline
    \end{tabularx}
    
    \label{ablation study}
\end{table}


\section{Conclusion}
In this study, we introduce \textit{DARLC}, a novel algorithm for dual advancement of representation learning and clustering for SNIs. \textit{DARLC} features in its enhanced data augmentation technique, comprehensive and potent representation learning approach that integrates MIM, CL and clustering, as well as robust and adaptable clustering algorithm. These features collectively contribute to \textit{DARLC}'s superiority in both image representation learning and clustering, as evidenced by our extensive benchmarks over multiple real datasets and comprehensive ablation studies.     

\section*{Acknowledgments}

This work was supported by Excellent Young Scientist Fund of Wuhan City under Grant 21129040740. We also thank Xiaowen Zhang and Nisang Chen for their helps in plotting figures and participation in discussions.

\clearpage
\footnotesize
\bibliographystyle{ACM-Reference-Format}
\bibliography{sample-base}

\end{document}